\pdfoutput=1
\documentclass[11pt]{article}

\usepackage{ACL2023}

\usepackage{times}
\usepackage{latexsym}
\usepackage{multirow}
\usepackage{todonotes}
\usepackage{comment}
\usepackage[multiple]{footmisc}
\usepackage{multido}
\usepackage[
  round-mode = places,
  round-precision = 2,
  round-pad = false %% true by default
]{siunitx}

\usepackage[T1]{fontenc}
\usepackage[utf8]{inputenc}
\usepackage{todonotes}

\usepackage{microtype}
\usepackage{inconsolata}
\usepackage{algorithm}
\usepackage{algpseudocode}

\usepackage{pifont}
\usepackage[normalem]{ulem}
\usepackage{amsmath}
\title{RetinaQA: A Robust Knowledge Base Question Answering Model \\for both Answerable and Unanswerable Questions}

\author{First Author \\
  Affiliation / Address line 1 \\
  Affiliation / Address line 2 \\
  Affiliation / Address line 3 \\
  \texttt{email@domain} \\\And
  Second Author \\
  Affiliation / Address line 1 \\
  Affiliation / Address line 2 \\
  Affiliation / Address line 3 \\
  \texttt{email@domain} \\}
\author{Prayushi Faldu$^\dag$, Indrajit Bhattacharya$^\ddag$, Mausam$^\dag$ \\
$^\dag$Indian Institute of Technology Delhi, $^\ddag$TCS Research \\
prayushifaldu123@gmail.com, b.indrajit@tcs.com, mausam@cse.iitd.ac.in
\\
\vspace{5cm}
}

\begin{document}
\maketitle
\newcommand{\olddata}{{ GrailQA}}
\newcommand{\data}{{ GrailQAbility}}
\newcommand{\sys}{{RetinaQA}}

\begin{abstract}

An essential requirement for a real-world Knowledge Base Question Answering (KBQA) system is the ability to detect answerability of questions when generating logical forms. 
However, state-of-the-art KBQA models assume all questions to be answerable. 
Recent research has found that such models, when superficially adapted to detect answerability, struggle to satisfactorily identify the different categories of unanswerable questions, and simultaneously preserve good performance for answerable questions. Towards addressing this issue, we propose \sys{}, a new KBQA model that unifies two key ideas in a single KBQA architecture: (a) discrimination over candidate logical forms, rather than generating these, for handling schema-related unanswerability, and (b) sketch-filling-based construction of candidate logical forms for handling data-related unaswerability. Our results show that \sys{} significantly outperforms adaptations of state-of-the-art KBQA models in handling both answerable and unanswerable questions and demonstrates robustness across all categories of unanswerability. 
Notably, \sys{} also sets a new state-of-the-art for answerable KBQA, surpassing existing models.
We release our code-base\footnote{\url{https://github.com/dair-iitd/RetinaQA}} for further research.

\end{abstract}
\section{Introduction}\label{sec:intro}

%There has been significant recent development in Knowledge Base Question Answering (KBQA) 
Question answering over knowledge bases (KBQA) 
%has received a lot of interest in recent years
~\cite{saxena:acl2022,zhang:acl2022subgraph,mitra:naacl2022cmhopkgqa,wang:naacl2022mhopkgqa,das:icml2022subgraphcbr,cao:acl2022progxfer,ye:acl2022rngkbqa,chen:acl2021retrack,das:emnlp2021cbr,shu-etal-2022-tiara,gu-etal-2023-dont} 
%where natural language questions are answered over a structured knowledge base, 
requires answering natural language questions over a knowledge base (KB),
%either via directly retrieving the answers~\cite{saxena:acl2022,zhang:acl2022subgraph,mitra:naacl2022cmhopkgqa,wang:naacl2022mhopkgqa,das:icml2022subgraphcbr}, or 
most commonly via generating formal queries or logical forms that are then executed over the knowledge base to retrieve the answers.
%~\cite{cao:acl2022progxfer,ye:acl2022rngkbqa,chen:acl2021retrack,das:emnlp2021cbr,shu-etal-2022-tiara,gu-etal-2023-dont}.
%The state-of-the-art models have multi-stage architectures starting with entity linking, retrieval of relevant KB data paths and schema elements (entity types and relations), and then producing logical forms using these, mostly via generation~\cite{ye:acl2022rngkbqa,chen:acl2021retrack,shu-etal-2022-tiara} or more recently via step-by-step discrimination~\cite{gu-etal-2023-dont}.
%While most approaches for producing logical forms are generative, very recently discriminative approaches have been proposed.
%Large and diverse benchmark datasets also exist for the task.
When users interact with KBs in real-world settings, unanswerability of questions arises naturally. 
Users are typically unfamiliar with the schema and data of the underlying KB. 
Further, KBs are also often incomplete.
While specialized models for handling unanswerability have been proposed for other question answering tasks~\cite{rajpurkar:acl2018-squadidk,choi:emnlp2018-quac,reddy:tacl2019coqa,sulem:naacl2022-ynidk,raina:acl2022-mcqnone},
all existing models for KBQA assume answerability of questions over the given KB.
%This is an unrealistic assumption since user queries are typically agnostic of the underlying KB, which is often incomplete.
%While specialized models for handling unanswerability have been proposed for other question answering tasks~\cite{rajpurkar:acl2018-squadidk,choi:emnlp2018-quac,reddy:tacl2019coqa,sulem:naacl2022-ynidk,raina:acl2022-mcqnone}, there is no such model for KBQA.

Recently,
\newcite{patidar-etal-2023-knowledge} published a 
%A recent study proposed a 
benchmark dataset called GrailQAbility 
%adapting the GrailQA dataset~\cite{gu:www2021grailqa} to 
incorporating different categories of unanswerable questions. 
%They also proposed the task of detecting unanswerabilty while answering KB questions.
This work also demonstrated that state-of-the-art KBQA models perform poorly off-the-shelf for unanswerable questions.
This performance improves with superficial adaptations for unaswerability, such as adding unanswerable questions during training and thresholding. 
%to separate the two question categories.
However, %this is shown to come at a cost.
such adaptations significantly hurt performance for answerable questions.
Additionally, different state-of-the-art models struggle with different categories of unaswerability, such as (a) questions for which schema elements (i.e. relations or entity types) are missing in the KB, and which therefore do not have valid logical forms, and
(b) questions for which data elements (i.e. entities or facts) are missing in the KB, and which therefore have logical forms that are valid, but return empty answers on execution.
This work demonstrated that there is no single model which performs well for all categories of unanswerable questions and answerable ones.
%This highlights the importance of rethinking KBQA architectures that are robust against different categories of unanswerability.

Our analysis of existing models reveals two key drawbacks that make KBQA models less robust -- imperfect calibration and over-reliance on path-based retrieval. 
To overcome these, we propose a new multi-stage \textbf{Ret}r\textbf{i}eve, ge\textbf{n}erate, and r\textbf{a}nk architecture for KBQA named \textbf{\sys{}}, which brings together several ideas from the literature.

First, a robust KBQA model has to separate questions that are answerable (having a valid logical form) from those that are unanswerable due to missing schema elements (not having a valid logical form). This requires good model calibration. %Unfortunately, 
Most KBQA models use \emph{generation} of logical form as their final step, which has calibration challanges. 
Instead, since discriminative models are generally better calibrated, \sys{}’s final stage \emph{discriminates} among candidate logical forms, %therefore 
%–  and this gives \sys{} 
% has a stronger ability to 
better separating questions without any logical form.

Secondly, existing models assume answerability, where all logical forms have accompanying
paths in the KB. These models learn to rely on
retrieved paths from the KB to generate logical
forms. However, missing data elements may break
any valid path, while the question, even though
unanswerable, still has a valid logical form. Existing models falter here, and end up generating
some other logical form corresponding to paths that
do exist in the KB. To handle such unanswerable
questions, RetinaQA additionally employs sketchfilling-based construction. It first generates schemaindependent high-level sketches, and then grounds
these for the specific KB using relevant schema
elements retrieved for the question.

Using experiments over GrailQAbility, we demonstrate that the performance of \sys{} 
%significantly outperforms 
is not only stable but also significantly better than that of 
adaptations of multiple state-of-the-art KBQA models that assume answerability, not only across different categories of unanswerable questions, but also for answerable ones.
Interestingly, we demonstrate that 
the \sys{} architecture performs strongly for fully answerable KBQA benchmarks as well, and establishes a new state-of-the-art performance on the GrailQA dataset.
%these architectural choices help for answerable questions as well.
%Discriminative scoring of syntactically and semantically valid candidates leads to clearer separation between correct and incorrect logical forms.
%We also demonstrate the advantages of 
%\sys{} for KBQA in general by outperforming existing KBQA models to establish a new state-of-the-art on the GrailQA dataset.

\begin{figure*}[t]
\centering
\includegraphics[width=0.95\textwidth]{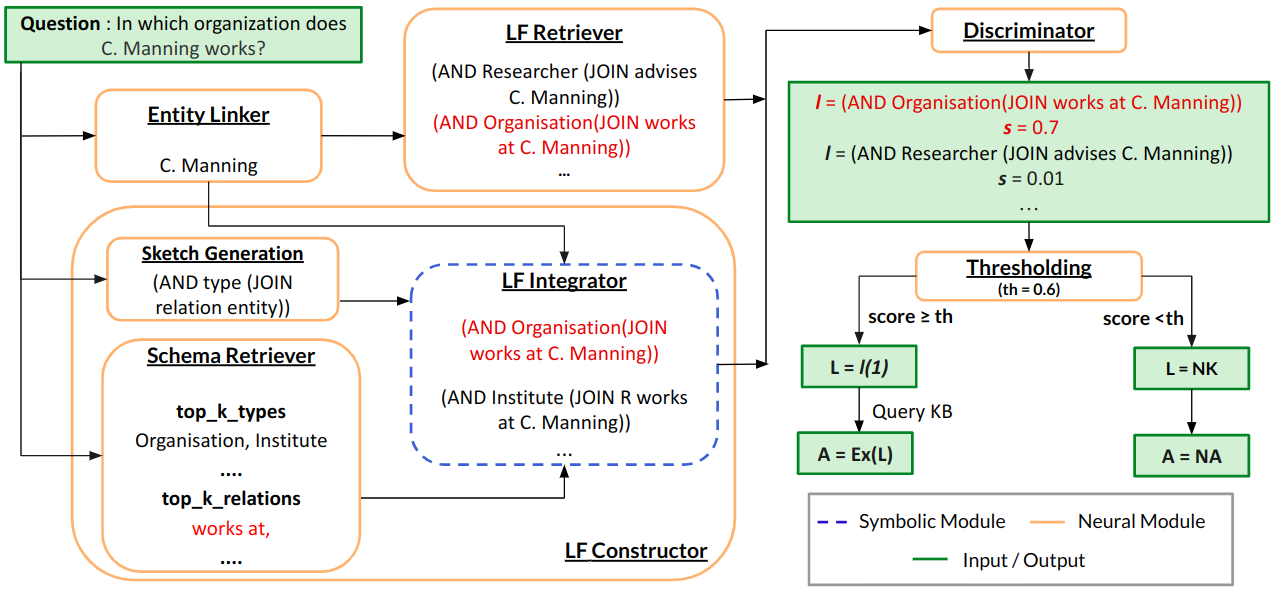}  
\caption{
%Model Architecture. Here $lf$ and $score$ represent candidate logical forms and scores assigned by the discriminator. $L$ refers to the final predicted logical form where $K$ is a valid logical form (lf with the highest score) and $NK$ refers to the logical form "Not Known". $A$ refers to the set of answers returned by KB. Note that when the predicted logical form is $NK$ the default answer is assigned default value i.e. NA.
\sys{} Architecture showing different components illustrated with an example question. Symbols $l$ and $s$ represent candidate logical form and its score as Discriminator output, $L$ the output logical form, $A$ the final answer, $l(1)$ the top ranked logical form, Ex($l$) the answer obtained by executing logical form $l$. NK and NA are special symbols indicating No Knowledge (for logical form) and No Answer. The logical form in red under LF Retriever would not be found if data element {\em (C. Manning, works at, Stanford)} is missing in the KB, and additionally that in red under LF Integrator would not be found if relation {\em works at} is missing in KB schema and therefore not retrieved by the Schema Retriever. The candidate logical form in red under Discriminator would not be found if both of these are missing.
}
\label{fig:example}
%\vspace{-0.5cm}
\end{figure*}

\section{Related Work}\label{sec:rw}

%There has been extensive recent research on Knowledge Base Question Answering.
%Though some models directly retrieve answers~\cite{saxena:acl2022,zhang:acl2022subgraph,mitra:naacl2022cmhopkgqa,wang:naacl2022mhopkgqa,das:icml2022subgraphcbr}, 
The predominant approach for supervised KBQA uses the question to construct a logical form, which is then executed to retrieve the answer~\cite{cao:acl2022progxfer,ye:acl2022rngkbqa,chen:acl2021retrack,das:emnlp2021cbr}. 
State-of-the-art models use $k$-hop path traversal to retrieve data paths from linked entities in the question~\cite{ye:acl2022rngkbqa,shu-etal-2022-tiara}. 
Some models, instead~\cite{chen:acl2021retrack} or additionally~\cite{shu-etal-2022-tiara}, retrieve schema elements (namely, entity types and relations) based on the question.
 These utilize transformer-based decoders to generate the target logical forms.
 % and employ constrained decoding during inference.
We have observed that generative models are not adequately calibrated to separate answerable and unanswerable questions correctly.

In contrast, 
%to this generative style, 
Pangu~\cite{gu-etal-2023-dont} uses language models to incrementally enumerate and discriminate between partial logical forms. Though this approach is better calibrated, its path-level enumeration makes it brittle for questions with missing data elements. Some retrieval-based methods ~\cite{saxena-etal-2020-embedkgqa, saxena:acl2022} also perform ranking of answer paths, but directly select answer nodes. These methods maximize the similarity score between a relation 
%(partial answer path) 
and the question. %Also, 
Adapting such techniques to detect unanswerable questions is difficult. 
%it will be difficult to adapt this type of technique .
In contrast, we perform contrastive-learning-based one-shot discrimination on fully-formed logical form candidates in the final stage.
% {\em We are not aware of any approach that performs one-shot discrimination on fully-formed logical form candidates as the final stage.}

In addition to supervised in-domain settings, transfer~\cite{cao:acl2022progxfer,ravishankar:emnlp2022} and few-shot~\cite{li-etal-2023-shot} settings have also been studied for KBQA.
Here, test questions involve unseen KB relations and entity types. %unseen during training.
Approaches for these settings first generate high-level sketches (also called drafts or skeletons) that capture the syntax of the target language. 
%but omit KB-specific arguments. 
These sketches are then filled in with KB-specific arguments to construct complete programs, which are finally scored and ranked.
Notably, these transfer architectures do not involve any traversal-based component to retrieve logical forms.
In contemporaneous work on few-shot transfer, ~\newcite{patidar-etal-2024-fusic} propose the FuSIC-KBQA framework accommodating one or more supervised retrievers coupled with an LLM for retrieval reranking and logical form generation. 
This demonstrates how retrievers from existing supervised models~\cite{shu-etal-2022-tiara,gu-etal-2023-dont} can be adapted and augmented using LLMs for low-supervision settings. 
In principle, \sys{} can also be accommodated as a retriever in this framework. 
%to boost its performance.

Unanswerability and specialized models for detecting unanswerable questions have been studied for many question-answering tasks~\cite{rajpurkar:acl2018-squadidk,choi:emnlp2018-quac,reddy:tacl2019coqa,sulem:naacl2022-ynidk,raina:acl2022-mcqnone}. 
%such as extractive QA~\cite{rajpurkar:acl2018-squadidk}, conversational QA~\cite{choi:emnlp2018-quac,reddy:tacl2019coqa}, Boolean (Y/N) QA~\cite{sulem:naacl2022-ynidk} and answering MCQs~\cite{raina:acl2022-mcqnone}.
However, no specialized models have been proposed for detecting unanswerable questions in KBQA.
All existing KBQA models assume that questions have valid logical forms with non-empty answers.
%and almost all existing datasets~\cite{gu:www2021grailqa,yih:acl2016webqsp,talmor:acl2018complexwebqsp,cao:acl2022kqapro} contain only answerable questions.
Even in the transfer setting for KBQA~\cite{cao:acl2022progxfer,ravishankar:emnlp2022}, though the target logical forms may involve schema elements unseen during training, questions are still assumed to be answerable.
Recent work~\cite{patidar-etal-2023-knowledge} has published the GrailQAbility benchmark by modifying the popular GrailQA dataset~\cite{gu:www2021grailqa} to incorporate multiple categories of unanswerable questions.
%This work also demonstrates the shortcomings of superficial adaptions of existing answerable-only KBQA models that assume answerability for detecting unanswerable questions.
This work also demonstrates that answerable-only KBQA models, when superficially adapted for handling unanswerability, fall short in many ways.

\section{The Problem and Our Solution}
\label{sec:model}

We first briefly define the KBQA with unanswerability task, and then describe the architecture of our proposed \sys{} model. %which we name \sys{}.

%\subsection{Overview}
%We propose a multi-staged method with alternating neural and symbolic stages which adds complementary strengths and makes the model robust against various types of unanswerabilites.

\subsection{KBQA with Unanswerability}

A Knowledge Base $G$ consists of a schema $S$ with data $D$ stored under it.
The schema consists of entity types $T$ and binary relations $R$ defined over pairs of types. 
Together, we refer to these as schema elements.
The data $D$ consists of entities $E$ as instances of types $T$, and facts $F\subseteq E \times R \times E$.
Together, we refer to these as data elements.
We follow the definition of ~\cite{patidar-etal-2023-knowledge} for defining the task of Knowledge Base Question Answering (KBQA) with unanswerability.
A natural language question $q$ is said to be answerable for a KB $G$ if it has a corresponding logical form $l$, which, when executed over $G$, returns a non-empty answer $A$.
In contrast, a question $q$ is unanswerable for $G$, if it either (a) does not have a corresponding logical form that is valid for $G$, or (b) it has a valid logical form $l$ for $G$, but which on executing returns an empty answer.
The first case indicates that $G$ is missing some schema element necessary for capturing the semantics for $q$.
The second case indicates that the schema $S$ is sufficient for $q$, but $G$ is missing some necessary data elements for answering it.
In the KBQA with unanswerability task, given a question $q$, if it is answerable, the model needs to output the corresponding logical form $l$ and the non-empty answer $A$ entailed by it, and if it is unanswerable, the model either needs to output NK (meaning No Knowledge) for the logical form, or a valid logical form $l$ with NA (meaning No Answer) as the answer. 
While different formalisms have been proposed for logical forms, we use {\em s-expressions}~\cite{gu:www2021grailqa}. 
These have set-based semantics, functions with arguments and return values as sets.
%These can be easily translated to KB query languages such as SPARQL, and provide a balance between readability and compactness~\cite{gu:www2021grailqa}.

%A Knowledge base stores knowledge in the form of RDF triples i.e., (s, r, o) where s is an entity, r is a relation, and o can be an entity or literal or type. Every entity is associated with an entity type. Let $E$ be set of all entities, $L$ be set of all literals, $T$ be set of all types and $R$ be set of relations. We refer to elements of $T$ and $R$ as schema elements while the entities and facts are referred to as data elements.  Let $F$ be a set of facts defined as $F \subset E \times R \times (E \cup L)$.
%We formulate the problem as a semantic parsing task where given a natural language question, the task is to convert it into an executable logical program that can be executed over KB to get answers. Questions that cannot attain a valid logical form are represented as $NK$.

\subsection{The \sys{} Model}
Fig.~\ref{fig:example} shows the architecture of \sys{}.
At a high level, \sys{} has two stages -- {\bf logical form enumeration}, followed by {\bf logical form ranking}.
For logical form enumeration, \sys{} follows two complementary approaches and then takes the union.
The first is {\bf path-traversal based retrieval}.
Starting from linked entities in the question, \sys{} traverses {\em data-level KB paths} and transforms these to logical forms. %\textcolor{blue}{i.e. this method utilize data-level knowledge of KB}.
The second is {\bf sketch-filling based construction}, which is critical when the KB has missing data elements for the question.
Here, \sys{} first {\em generates} {\bf logical form sketches} corresponding to the question, and then {\em enumerates} semantically valid groundings for these by {\em retrieving} relevant KB schema elements for filling in the sketch arguments. 
Note that this approach utilizes only the KB schema and avoids data.
%This construction eliminates all invalid candidates using symbolic checking for syntactic (i.e. logical form grammar) and semantic (i.e. schema-level constraint) correctness.
Once candidate logical forms are so identified, \sys{} uses {\em discriminative} scoring to rank these logical forms with respect to the question.
%We prefer a discriminative approach over the generative approach of most existing KBQA approaches, since the former results in better calibration, which in turn leads to more accurate thresholding.
%However, one requirement for a discriminative approach is limiting the space of candidates so that it contains as few irrelevant ones as possible.
We next explain each of these components in more detail.

%Here we propose a unified KGQA model that is robust against different types of unanaswerabilities with minimal effect on the performance of the answerable questions. GrailQAbility ~\cite{patidar-etal-2023-knowledge} presents an extensive study on various KGQA methods and how different modeling choices affect different types of unanaswerabilities. Here we highlight the key intuitions from the ~\cite{patidar-etal-2023-knowledge} which motivates the architecture of the proposed model:
%\begin{enumerate}
%  \item Logical Form thresholding helps to improve the performance of schema level unanswerabilites.
%  \item KG-dependent enumeration-based methods limits the performance of data level unanswerabilites while boost performance of zshot answerable questions. 
%  \item KG-independent retriever-based methods helps to improve the performance of data level unanswerabilites while they are not good for zshot answerable questions.
%  \item Generation-based models trained to generate programs are not very well calibrated, which affects the performance of answerable questions after A+U training.
%  \item Interactions with KG is very limited and is performed only during inference after training. 
%\end{enumerate}

%\subsection{Retriever}

\paragraph{Entity Linker:} The pipeline starts with linking mentioned entities in the question with KB entities $E$. 
This is required for both logical form retrieval and logical form construction.
We use an off-the-shelf entity linker~\cite{ye:acl2022rngkbqa} previously used in the KBQA literature~\cite{shu-etal-2022-tiara,gu-etal-2023-dont}. 
%which uses a standard 3-staged pipeline - Mention Detection, Candidate Generation, and Entity Disambiguation.  
More details are in the Appendix (\ref{subsec:appendix_entity_linker}).
If the mentioned entities are missing in the KB, the entity linker returns an empty set.

\paragraph{Logical Form Retriever: }
As the first approach for enumerating logical forms, \sys{} uses KB data path traversal~\cite{ye:acl2022rngkbqa}.
\sys{} traverses 2-hop paths starting from the linked entities and transforms these to logical forms in s-expression. 
These logical forms are then scored according to their similarity with the question,
%, ranked according to these scores 
and the top-10 logical forms are selected for the next stage, as illustrated under LF Retriever in Fig.~\ref{fig:example}.
Following~\citep{ye:acl2022rngkbqa}, we score a logical form $l$ and question $q$ as:
\begin{equation}
    s(l,q) = \textsc{Linear}(\textsc{Bert}\textsc{Cls}([l;q]))
\label{eqn:lf_scorer}
\end{equation}
and optimize a contrastive objective for ranking:
\begin{equation}
    \mathcal{L}_{ret} = - \frac{\exp(s(l^*,q))}{\exp(s(l^*,q)) + \sum_{l\in L\land l\neq l^*}\exp(s(l,q))}
\label{eqn:retrieval_loss}
\end{equation}
where $l^*$ is the gold-standard logical form for $q$, and $L$ is the set of logical forms similar to $l^*$.
Note that the transformation to logical forms from KB-paths only covers certain operators (such as {\em count}), but not some others (such as {\em argmin}, {\em argmax}), so that this enumeration approach is not guaranteed to cover all logical forms.
More importantly for unanswerability, as illustrated in Fig.~\ref{fig:example}, this approach cannot retrieve the logical form in red when the relevant data path in the KB is broken, as by the absence of the data element {\em (C. Manning, works at, Stanford)} in our example.
%Note also that this approach cannot retrieve the correct logical form when $q$ has a valid logical form, but no connected data path leading to an answer. 

%\textbf{Logical Form Retriever}: We follow the Enumerate and Rank method as proposed by~\cite{ye:acl2022rngkbqa}. It starts with entity mentions and enumerates all m-hop paths. Since enumeration grows exponentially with number of hops, we enumerate only 2-hop paths i.e. m=2. Each path is deterministic-ally converted into a logical form i.e. s-expression. All possible candidate logical forms are fed into a Ranker with questions that assign a score to every question, logical form pair. We select top-10 logical forms from the candidates based on the score. The Enumerate and Rank method does not guarantee to cover the target logical form. 
%\\\\

\paragraph{Logical Form Constructor:} 
The second approach used by \sys{} for logical form enumeration is sketch-filling.
Drawing inspiration from the transfer approaches for KBQA~\cite{cao:acl2022progxfer,ravishankar:emnlp2022,li-etal-2023-shot}, \sys{} uses logical form sketches. 
Sketches capture KB-independent syntax of s-expressions with functions, operators and literals, and replace KB-specific elements, specifically entities, entity types and relations, with arguments. 
\sys{} first generates sketches using a {\bf sketch generator}, and in parallel retrieves relevant schema elements as candidates for sketch arguments using a {\bf schema retriever}, and finally fills in arguments for each candidate sketch using the retrieved argument candidates in all possible valid ways using a {\bf logical form integrator}.
By avoiding path-based retrieval, this approach can construct valid logical forms when these exist, even when some relevant data element for the question is missing in the KB, for example, when the data element {\em (C. Manning, works at, Stanford)} is missing in the KB but the relation {\em works at} is present in the KB schema.
%\textcolor{blue}{For example in Figure~\ref{fig:example} if there is no edge "works at" from C. Manning then LF Retriever cannot enumerate logical form marked in red while Logical Form Integrator can do.}

\textbf{Sketch Generator:} 
The sketch generator takes the question $q$ as input and outputs a sketch $s$, optimizing a cross-entropy-based objective:
\begin{equation*}
    L_{sketch} = - \sum_{t=1}^{n} \log(p (s_t | s_{<t},q))
\end{equation*}
%where $q$ is question, $s$ is target sketch with $n$ tokens and $L_{sketch}$ is parsing loss.
Specifically, we fine-tune T5~\cite{raffel:jmlr2020} as the Seq2Seq model.
We also perform constrained decoding during inference to ensure syntactic correctness of the generated sketch. 
This step is unaffected by any KB incompleteness.

\textbf{Schema Retriever:}  
To retrieve candidate arguments for generated sketches, we follow the schema retriever of \textsc{TIARA}~\cite{shu-etal-2022-tiara}. 
%(Note that \textsc{TIARA} does not have a sketch generator, but instead uses retrieved schema elements as input to a logical form generator.)
% It works very similarly to the logical form retriever, but with schema elements instead of logical forms.
It is a cross encoder and uses the form of Eqn.\ref{eqn:lf_scorer} to score a schema element $x$ and the question $q$, and optimizes the same form of the objective as for the sentence-pair classification task ~\cite{devlin-etal-2019-bert}.
We train two retriever models, one for relations and one for types, and use the top-10 types and top-10 relations as candidate arguments for each question.
As illustrated in Fig.~\ref{fig:example}, this step breaks when relevant relations, such as {\em works at}, or entity types are missing from the KB schema.

\textbf{Logical Form Integrator:} 
This component grounds each generated candidate sketch using the retrieved candidate arguments and the linked entities to construct complete logical form candidates.
Each candidate sketch is grounded using every possible combination of candidate arguments.
A symbolic checker ensures type-level validity of the grounded logical forms for the KB $G$. 
This also avoids a combinatorial blow-up and restricts the space of logical form candidates.
This component does not involve any trainable parameters.

\paragraph{Logical Form Discriminator:} 
This component finally scores and ranks all the logical form candidates provided by the retriever and constructor components.
A T5 encoder-decoder model is trained to score a logical form candidate. 
Following ~\newcite{zhuang2022rankt5}, we feed a (question, logical form) pair to the encoder, and use decoding probability for a special token as the ranking score.\footnote{We use $<extra\_id\_6>$ token of T5 for tuning the ranking score.}
This component uses a contrastive learning-based optimization objective similar to Eqn.\ref{eqn:retrieval_loss}.
We sample negative examples randomly, but this mostly covers the set of all negative candidates given its small size.
%hence for most of the questions negative samples cover the entire negative candidate set.

For a test question, the candidate logical forms are ranked according to the predicted discriminator scores. 
If the score of the top-ranked candidate is below a threshold (tuned on the validation set), it is classified as unanswerable i.e. $L=$ NK and $A=$ NA. 
This helps in identifying questions for which valid logical forms do not exist due to missing schema elements.
For example, in Fig.~\ref{fig:example}, if the logical form in red is missing from the candidate list, the discriminator assigns a low score to the rank 1 logical form candidate, and NK is output after thresholding.
Otherwise, the top-ranked candidate is predicted as the logical form. 

The predicted logical form is then converted to SPARQL and executed over KB. 
If a non-empty answer is obtained, then the question is considered answerable with the output as the desired answer.
On the other hand, if the answer is empty, the question is classified as unanswerable under the missing data elements category, i.e., $A=$ NA.

\section{Experiments} \label{sec:experiments}
% We address the following research questions: (1) How does \sys{} compare against existing KBQA approaches, in  settings that have both answerable and unanswerable questions, and also in those that have only answerable questions? 
% (2) How does \sys{} perform for different categories of unanswerable questions, i.e., those that are unanswerable due to missing schema elements, and those with missing data elements? 
% (3) What are the individual contributions of various model components in towards the performance of \sys{} in (1) and (2) above? 
% \textcolor{blue}{(4) How does \sys{} compare against existing KBQA approaches in answerable-only settings?}
We address the following research questions: {\bf (1)} How does \sys{} compare against baselines for answerable and unanswerable questions in  two different training settings: one with only answerable questions and another with both answerable and unanswerable questions?
% (2) How does \sys{} perform for different categories of unanswerable questions, i.e., those that are unanswerable due to missing schema elements, and those with missing data elements? 
{\bf (2)} How does \sys{} perform for different categories of unanswerable questions? 
{\bf (3)} How does \sys{} perform in the answerable-only setting?
{\bf (4)} To what extent do the different components of \sys{} contribute towards its performance in (1), (2) and (3) above?

\begin{table*}[h!]
\begin{center}
\small
%\resizebox{1\linewidth}{!}{
            \begin{tabular}[b]{|c|l|rrr|rrr|rrr|}
%\toprule
%\textbf{Memory} & \textbf{Approach} & \multicolumn{5}{c}{\textbf{Datasets}} \\ 
\hline

  \multicolumn{1}{|c|}{Train}&\multicolumn{1}{c|}{Model}&\multicolumn{3}{c|}{Overall}
  & \multicolumn{3}{c|}{Answerable}  & \multicolumn{3}{c|}{Unanswerable}  \\

 & & F1(L) & F1(R) & EM & F1(L) & F1(R) & EM & F1(L) & F1(R) & EM \\ \hline
  \multirow{8}{*}{A}& RnG-KBQA &	67.80 &	65.60 &	51.60 &	78.10 &	78.10 &	74.20 &	46.90 &	40.10 &	5.70 \\
& RnG-KBQA + T &	67.60 &	65.80 &	57.00 &	71.40 &	71.30 &	68.50 &	59.90 &	54.50 &	33.60 \\
& Tiara &	75.05 &	72.84 &	53.69 &	80.03 &	80.00 &	75.63 &	64.95 &	58.31 &	9.20 \\
& Tiara + T &	73.26 &	71.62 &	55.23 &	74.08 &	74.05 &	70.56 &	71.60 &	66.68 &	24.15 \\
& Pangu & 63.09 & 60.06 & 54.55 & 78.72 & 78.7 & 74.00 & 31.40 & 22.25 & 15.13 \\
& Pangu + T & 79.14 & 77.89 & 66.53 & 75.52 & 75.51 & 72.37 & 86.48 & 82.70 & 54.68 \\
& \sys{} &	76.83 &	75.24 &	64.54 &	{\bf81.22} &	{\bf81.2} &	{\bf77.41} &	67.93 &	63.16 &	38.45 \\
& \sys{} + T &	{\bf81.80} &	{\bf80.78} &	{\bf73.76} &	77.74 &	77.74 &	75.01 &	{\bf90.02} &	{\bf86.96} &	{\bf71.20} \\
  \hline 
  \multirow{11}{*}{A+U}& RnG-KBQA &	80.50 &	79.40 &	68.20 &	75.90 &	75.90 &	72.60 &	89.70 &	86.40 &	59.40 \\
& RnG-KBQA + T &	77.80 &	77.10 &	67.80 &	70.90 &	70.80 &	68.10 &	92.00 &	89.80 &	67.20 \\
& Tiara &	78.29 &	77.43 &	66.29 &	71.33 &	71.32 &	68.29 &	92.4 &	89.82 &	62.24 \\
& Tiara + T &	77.67 &	76.94 &	66.87 &	69.89 &	69.88 &	66.98 &	93.43 &	91.24 &	66.65 \\
& Pangu &	63.59 &	60.42 &	53.79 &	79.45 &	79.42 &	73.49 &	31.42 &	21.89 &	13.85 \\
& Pangu + T &	78.29 &	76.91 &	66.14 &	75.25 &	75.22 &	71.62 & 80.46 &	80.32 &	55.03 \\
& \sys{} &	77.31 &	75.71 &	64.79 &	{\bf80.98} &	{\bf80.97} &	{\bf76.95} &	69.87 &	65.04 &	40.14 \\
& \sys{} + T &	{\bf83.30} &	{\bf82.69} &	{\bf77.45} &	77.91 &	77.91 &	75.16 &	{\bf94.21} &	{\bf92.38} &	{\bf82.10} \\
\cline{2-11}
 %  \hline 
 % \multirow{3}{*}{A+U}
& \sys{} - LFR + T &	77.36 & 76.37 & 65.37 & 73.40 & 73.39 & 70.90 & 85.38 & 82.43 & 54.17 \\ 
& \sys{} - LFI + T &	74.89 & 73.53 & 53.89 & 70.89 & 70.85 & 68.07 & 83.01 & 78.95 & 25.13 \\ 
& \sys{} - (SG $\cup$ SR) + T & 64.68 & 62.58 & 52.46 & 72.99 & 72.95 & 68.13 & 47.84 & 41.54 & 20.70 \\ 
\hline 
\end{tabular}
%}
\end{center}
\caption{Performance of different models on the GrailQAbility dataset: overall and for answerable and unanswerable questions. A indicates training with answerable questions, A+U with answerable and unanswerable questions, +T indicates thresholding.
%EM is exact match on logical forms and F1(L) and F1(R) are lenient and regular evaluations of answers. A and A+U indicate training with only answerable questions and with both answerable and unanswerable questions. Models with suffix +T have additional thresholds for entity disambiguation and logical form fine-tuned on dev set.
Ablations of \sys{} are named as \sys{} - X, where we denote logical form retriever as LFR, logical form integrator as LFI and sketch generator and schema retriever together as (SG $\cup$ SR).
}
\label{tab:mainresult}
%\vspace{-0.5cm}
\end{table*}

\begin{table*}[h!]
        \begin{center}
            \small
            % \scriptsize
            %\resizebox{1\linewidth}{!}{
            \begin{tabular}[b]{|c|l|rrrrrr|rrrr|}
%\toprule
%\textbf{Memory} & \textbf{Approach} & \multicolumn{5}{c}{\textbf{Datasets}} \\ 
\hline
  \multicolumn{1}{|c|}{Train}&\multicolumn{1}{c|}{Model}&\multicolumn{6}{c|}{Schema Element Missing}  
  & \multicolumn{4}{c|}{Data Element Missing}  \\     
  & &\multicolumn{2}{c}{Type}  
  & \multicolumn{2}{c}{Relation}  & \multicolumn{2}{c|}{Mention Entity} &
  \multicolumn{2}{c}{Other Entity} &
  \multicolumn{2}{c|}{Fact}  \\     
  & & F1(R) & EM &  F1(R) & EM  & F1(R) & EM &F1(R) & EM & F1(R) & EM  \\ \hline
%  \multirow{8}{*}{A}& RnG-KBQA &	40.1 &	0 &	44.2 &	0 &	27.4 &	0 &	45.1 &	13.5` &	46 &	16.8 \\
\multirow{4}{*}{A}& RnG-KBQA + T &	55.50 &	49.50 &	57.10 &	46.60 &	44.70 &	40.30 &	56.00 &	11.50 &	58.60 &	13.90 \\
%& Tiara &	59.55 &	0.71 &	58.07 &	0 &	50.63 &	0 &	64.15 &	23.53 &	62.71 &	26.01 \\
& Tiara + T &	66.27 &	21.70 &	70.21 &	28.06 &	61.01 &	23.43 &	68.91 &	22.97 &	68.29 &	23.63 \\
%& Pangu & 24.06 & 19.46 & 21.13 & 16.14 & 41.67 & 38.84 & 12.89 & 0 & 13.06 & 0.95 \\
& Pangu + T & {\bf87.97} & {\bf87.50} & {\bf80.07} & {\bf79.63} & 90.57 & {\bf90.41} & 83.19 & 0.00 & 76.48 & 1.07 \\
%& \sys &	45.4 &	19.34 &	47.45 &	7.15 &	65.57 &	41.04 &	{\bf85.71} &	{\bf68.91} &	{\bf84.68} &	{\bf72.45} \\
& \sys{} + T &	86.32 &	80.31 &	79.41 &	62.08 &	{\bf90.72} &	77.83 &	{\bf 91.60} &	{\bf 68.07} &	{\bf 93.11} &	{\bf 71.14} \\
  \hline 
%  \multirow{8}{*}{A+U}& RnG-KBQA &	91.6 &	75.8 &	86.4 &	66.6 &	87.6 &	72 &	84 &	37.5 &	82.4 &	39.1 \\
\multirow{4}{*}{A+U} & RnG-KBQA + T &	93.40 &	86.80 &	89.70 &	{\bf85.50} &	92.10 &	89.60 &	87.10 &	30.80 &	86.00 &	32.50 \\
%& Tiara &	90.09 &	76.53 &	88.95 &	62.62 &	92.14 &	63.68 &	91.32 &	55.18 &	90.14 &	54.99 \\
& Tiara + T &	91.63 &	83.84 &	{\bf90.90} &	72.37 &	{\bf94.50} &	71.38 &	91.60 &	50.42 &	90.38 &	52.85 \\
%& Pangu &	25.24 &	17.57 &	19.72 &	13.65 &	42.45 &	37.42 &	11.76 &	0 &	12.35 &	0.71 \\
& Pangu + T &	90.80 &	{\bf 90.68} &	78.66 &	78.44 &	90.41 &	{\bf90.25} &	74.51 &	0.00 &	69.71 &	0.95 \\
%& \sys &	49.65 &	19.22 &	49.08 &	7.48 &	65.09 &	40.88 &	89.64 &	75.91 &	86.1 &	75.89 \\
& \sys{} + T &	{\bf94.22} &	90.21 &	88.52 &	81.91 &	94.34 &	86.64 &	{\bf93.84} &	{\bf75.91} &	{\bf94.30} &	{\bf76.13} \\
  \hline 
\end{tabular}
%}
\end{center}
\caption{Performance of different models for the unanswerable questions in GrailQAbility, grouped by categories of KB incompleteness. 
Note that missing mention entities result in invalid logical form, while other missing entities lead to valid logical form with no answer. 
%Names have the same meanings as in Tab.~\ref{tab:mainresult}.
}
\label{tab:drop_type}
%\vspace{-0.5cm}
\end{table*}

\subsection{Experimental Setup}
%\textcolor{blue}{[Should we highlight here that primary goal of paper is to build a robust model and there is only one dataset for that. However to show that method is competent enough for datasets with only "answerable" questions we have done experiments on two additional datasets???]}

\textbf{Datasets: } 
For research questions (1) and (2) above, we use the GrailQAbility dataset, which is the only KBQA dataset that contains both answerable and unanswerable questions.
For research question (3), we use the two most popular KBQA datasets with only answerable questions, namely GrailQA and WebQSP.
%We use GrailQA ~\cite{gu:www2021grailqa}, WebQSP~\cite{yih:acl2016webqsp} and GrailQAbility~\cite{patidar-etal-2023-knowledge} as our datasets.

{\bf GrailQA} ~\cite{gu:www2021grailqa} is a popular KBQA dataset that contains only answerable questions. 
% It has 64,331 questions and their associated logical forms. 
%The background KB is Freebase. 
It contains questions at various levels of generalization: IID (seen schema elements), compositional (unseen combination of seen schema elements) and zero-shot (unseen schema elements). 
% The most complex questions can have multiple operators and up to 4 relations.
%While GrailQA only contains answerable questions, 
{\bf WebQSP} ~\cite{yih:acl2016webqsp} also has only answerable questions with Freebase as the KB, but unlike GrailQA, where the questions are synthetically constructed, it contains real user queries annotated with logical forms. It only has IID test questions.
{\bf GrailQAbility} ~\cite{patidar-etal-2023-knowledge} is a recent dataset that adapts GrailQA to additionally incorporate unanswerable questions. The unanswerable questions are constructed by systematically dropping data and schema elements from the KB. More details are added in the Appendix (\ref{subsec:appendix_dataset_creation}).
%{\em This is the only existing KBQA dataset that contains unanswerable questions.} 

%\textbf{GrailQAbility} [cite] is the most diverse dataset of KGQA which contains both answerable and unnasnwerable questions. It is built on Freebase [cite] and has 50,507 questions annotated with logical forms. It addition to answerable questions across different levels of generalisations it also contains unanswerable questions with affected by different types of unanswerabilties.Trade-offs between different categories of answerable and unasnwerable questions makes dataset interesting to work with. We have updated the original Freebase KG with the steps as described in paper.
%Since our primary goal is to design a robust KGQA model which performs good for both answerable and unanswerable questions, we have also performed experiments with GrailQA which is largest dataset of KGQA with  64,331 questions annotated with logical forms. GrailQA contains only answerable questions but it has questions across various levels of generalisations.
%Both datasets have questions with differents operators and questions having up-to 4 relations.

\vspace{0.5ex}
\noindent
\textbf{Evaluation Metrics: } We primarily focus on evaluating the logical form using the Exact Match (EM) metric, which checks whether the predicted logical form is same as the gold logical form (which is NK for unanswerable questions with missing schema elements). 
We also evaluate the answers using the F1 score, which compares the predicted answer set with the gold answer set. 
For unanswerable questions, similar to prior work \cite{patidar-etal-2023-knowledge}, we report two F1 scores -- the regular score F1(R) compares the list of answers based on the given incomplete KB, 
% and the lenient score accepts an answer even if it is absent from the given KB, but was present in the original GrailQA KB. 
and the lenient score F1(L) which does not penalize a model for returning ideal answers. 
Specifically, F1(L) accepts answers considering both KBs -- the ideal or complete KB and new or incomplete KB. 
In a way, it evaluates the model's ability to infer missing KB elements and predict the correct answer.

%More details about the metrics are in the Appendix.

%GrailQAbility evauates logical forms using Exact Match (EM) and evaluates answers using F1 scores. GrailQAbility employs two different types of F1 scores i.e. F1 (L) and F1 (R). We focus more on EM as we expect a good KGQA model should also have good reasoning abilities.

\vspace{0.5ex}
\noindent
\textbf{Baselines: }
We compare \sys{} against existing state-of-the-art KBQA models, as per the GrailQA leaderboard and code availability. 
These are TIARA~\cite{shu-etal-2022-tiara}, RnG-KBQA ~\cite{ye:acl2022rngkbqa}, and Pangu ~\cite{gu-etal-2023-dont}. Of these, the first two are shown to the best performing models on GrailQAbility, and Pangu is a SoTA model for GrailQA\footnote{\url{https://dki-lab.github.io/GrailQA/}} and WebQSP~\cite{gu-etal-2023-dont}. 
For fair comparison, we use the same entity linker \cite{ye:acl2022rngkbqa} and T5-base as base LLM for all models.
For GrailQAbility, we adapt all models appropriately for unanswerability.
Specifically, we perform thresholding (denoted as "+ T") on 
%entity disambiguation and 
logical form generation to output NK. 
The thresholds are tuned on the dev set. 
%\textcolor{red}{For \sys{} and Pangu, the entity disambiguation threshold was negatively impacting overall performance. Therefore, we applied only the logical form threshold to these models.}
Additionally, instead of training the models only on answerable questions (denoted as A training), we train the models using both answerable and unanswerable questions (denoted as A+U training).
Further implementation details are in the Appendix (\ref{subsec:implementation_details}).

%We have selected baselines for comparision based on GrailQAbility experiments, GrailQA leaderboard and code availability during submission. Two best performing baselines for GrailQAbility ~\cite{patidar-etal-2023-knowledge} are TIARA ~\cite{shu-etal-2022-tiara} and RnG-KBQA ~\cite{ye:acl2022rngkbqa} . They are also among the top-3 published models (with code availability) on GrailQA leaderboard. In this work we add one more baseline Pangu ~\cite{gu-etal-2023-dont} which is current state-of-the-art KGQA model for GrailQA.
%For GrailQAbility we have re-run all baselines by making appropriate code changes to adapt it for answerability detection.
%To make fair comparison we have used same entity linker ~\cite{ye:acl2022rngkbqa} and same base models (T5-base) across all baselines. Details are discussed in appendix section.

\begin{table*}[h!]
\begin{center}
\small            %\resizebox{1\linewidth}{!}{
\begin{tabular}[b]{|c|l|rrr|rrr|rrr|}
%\toprule
%\textbf{Memory} & \textbf{Approach} & \multicolumn{5}{c}{\textbf{Datasets}} \\ 
\hline
  \multicolumn{1}{|c|}{Train}&\multicolumn{1}{c|}{Model}&\multicolumn{3}{c|}{IID}
  
  & \multicolumn{3}{c|}{Compositional }  & \multicolumn{3}{c|}{Zero-Shot}  \\

 & & F1(L) & F1(R) & EM & F1(L) & F1(R) & EM & F1(L) & F1(R) & EM \\ \hline
  \multirow{4}{*}{A}& RnG-KBQA &	85.50 &	85.40 &	83.20 &	65.90 &	65.90 &	60.20 &	72.70 &	72.70 &	67.30 \\
%& RnG-KBQA+T &	79 &	79 &	77.3 &	58.8 &	58.8 &	54.5 &	65.8 &	65.8 &	61.9 \\
& TIARA &	86.53 &	86.47 &	84.52 &	72.02 &	72.02 &	64.93 &	74.24 &	74.24 &	67.60 \\
%& TIARA + T &	83.76 &	83.71 &	82.01 &	65.68 &	65.68 &	60.86 &	64.02 &	64.02 &	58.58 \\
& Pangu & 82.00 & 81.97 & 79.09 & 71.63 & 71.63 & 65.95 & {\bf77.02} & {\bf77.02} & {\bf70.18} \\
%& Pangu + T & 81.15 & 81.14 & 78.88 & 68.69 & 68.69 & 64.73 & 70.46 & 70.46 & 66.41 \\
& \sys &	{\bf87.94} &	{\bf87.90} &	{\bf85.85} &	{\bf73.92} &	{\bf73.92} &	{\bf67.48} &	74.84 &	74.84 &	69.68 \\
%& \sys + T &	87.88 &	87.85 &	84.38 &	73.92 &	73.92 &	63.91 &	74.84 &	74.84 &	66.49 \\
  \hline 
  \multirow{4}{*}{A+U}& RnG-KBQA &	85.40 &	85.30 &	83.30 &	65.80 &	65.80 &	60.80 &	66.90 &	66.90 &	62.60 \\
%& RnG-KBQA+T &	80.9 &	80.9 &	79.2 &	60.5 &	60.5 &	56.1 &	61.1 &	61.1 &	57.6 \\
& TIARA &	82.38 &	82.36 &	80.57 &	65.16 &	65.16 &	59.84 &	58.50 &	58.50 &	54.65 \\
%& TIARA + T &	81.86 &	81.84 &	80.09 &	64.23 &	64.23 &	59.02 &	55.56 &	55.56 &	51.99 \\
& Pangu &	81.08 &	81.01 &	76.85 &	{\bf77.43} &	{\bf77.43} & {\bf69.52} &	{\bf78.01} &	{\bf78.01} &	{\bf70.42} \\
%& Pangu+T &	81.07 &	81 &	75.72 &	77.43 &	77.43 &	69.22 &	{\bf78} &	{\bf78} &	{\bf66.9} \\
& \sys &	{\bf89.00} &	{\bf88.98} & {\bf87.06} &	71.69 &	71.69 &	65.55 &	73.59 &	73.59 &	67.51 \\
%& \sys + T &	88.36 &	88.35 &	86.86 &	68.65 &	68.65 &	64.63 &	67.14 &	67.14 &	63.17 \\
  \hline 
\end{tabular}
%}
\end{center}
\caption{Performance of different models for answerable questions in the GrailQAbility dataset, for IID, compositional, and zero-shot test scenarios. Names have the same meanings as in Table~\ref{tab:mainresult}.
}
\label{tab:test_scenarios_answerable}
%\vspace{-0.5cm}
\end{table*}

\subsection{Results for KBQA with Unanswerability} \label{sec:results}

%We have compared \sys with various baselines for GrailQAbility(Tab.~\ref{tab:mainresult}) and GrailQA( Tab.~\ref{tab:grailqa_main_table}). For fair comparison we have used same entity linker pipeline for all the baselines (Details are explained in Implementation Section).

%\subsection{GrailQAbility}

We first address research question (1).
In Table \ref{tab:mainresult}, we report high-level performance of different models on GrailQAbility. 
%Similar to \citep{patidar-etal-2023-knowledge}, we divide the discussion into two settings: A+U training, where models are trained on both answerable and unanswerable questions, and A training, where they are trained on only the former.
With A+U training, \sys+T outperforms all models overall and is about 9 pct points  ahead of the closest competitor (Pangu+T) in terms of EM. 
For unanswerable questions, \sys{} achieves a 16 pct points improvement, while being consistently better for answerable questions. 
Unsurprisingly, thresholding helps all models for unanswerable questions and hurts slightly for answerable ones. 
This drop is relatively small for Pangu and \sys, suggesting that these are better calibrated due to their discriminative training.

Next, we address research question (2).
Table \ref{tab:drop_type} drills down on performance for different categories of unanswerability. 
First, we observe that for the baselines, performance varies significantly across different categories.
Pangu performs well for missing schema elements but is the worst model for missing data elements. 
TIARA is the best baseline for missing data elements but does not perform as well for missing schema elements. 
We further analyze such variability in performance for the baselines in the Appendix (Sec.~\ref{subsec:trade-off}).
\sys{} performs the best by a large margin for questions with missing data elements, and compares favorably with Pangu for missing schema elements, making it the overall model of choice across different categories of unanswerability. 
We also note that \sys{}+T results shows little degradation for questions with missing data (which have valid logical forms), and huge gains for questions with missing schema elements. 

%\textcolor{red}{[Why is this here and not under ablations?]} \textcolor{blue}{I think the below content seems reptitive in some sense. \sout{Our ablations (Section~\ref{subsec:ablations}) suggest that data drop gains in \sys{} are primarily due to its enumeration-independent schema retriever.
%We further compare logical form generation (RnG-KBQA, TIARA) and logical form discrimination based approaches (Pangu, \sys{}) on unanswerability due to missing schema elements. 
%We notice that for RnG-KBQA and TIARA the performance gain comes mainly from A+U training rather than from thresholding, leading to a drop in performance for answerable questions. 
%However, for Pangu and \sys{}, thresholding makes a significant contribution to the performance gain,  leading to robust performance for answerable questions.}}

%baselines are not consistently good. For eg Pangu is good at schema drop while is worst for data drop, while TIARA is best baseline for data drop but it is not as good as other baselines for schema drop. \sys\ is best for data drop while it has comparable performance for schema drop with previous SOTA. Again it can be observed here that schema drop questions gains a huge delta in performance with thresholding with a minimal or no loss for data drop questions. The gain for data drop is mainly due to schema retriever (enumeration independent), details are described in Section:~\ref{sec:analysis}.

In the A training setting, as a testimony to its robustness, \sys{} achieves comparable performance for answerable and unanswerable questions, with an EM gap of only 4 pct points.
In contrast, this gap ranges 18 to 45 pct points for other models. 
Other trends are very similar to the A+U setting. 
%Detailed Analysis is in Section ~\ref{sec:analysis}.

Additionally, we see that \sys{} largely outperforms existing models across different generalization settings for answerable questions (Table ~\ref{tab:test_scenarios_answerable}). 
\sys{} performs the best for IID and compositional generalization, but for zero-shot generalization, \sys{} performs slightly worse than Pangu. 
This is mainly because of the trade-off related to path traversal, as we explain in Section~\ref{subsec:ablations}. 

For unanswerable questions (Table ~\ref{tab:test_scenarios}), \sys{} outperforms all existing models for both IID and zero-shot generalization. Furthermore, we observe that the difference between EM and F1(R) scores is the smallest by far for \sys{}, indicating that it not only predicts unanswerability correctly but also does so for the right reason.

\begin{table}[h!]
\begin{center}
\small            %\resizebox{1\linewidth}{!}{
\begin{tabular}[b]{|c|cc|cc|}
%\toprule
%\textbf{Memory} & \textbf{Approach} & \multicolumn{5}{c}{\textbf{Datasets}} \\ 
\hline
  \multicolumn{1}{|c|}{Model}&\multicolumn{2}{c|}{IID}
  
  & \multicolumn{2}{c|}{Zero-Shot }    \\

 & F1(R) & EM & F1(R) & EM  \\ \hline
%  \multirow{3}{*}{A}&RnG-KBQA & & & &   \\ 
% &RnG-KBQA+T & & & &   \\ 
%  &ReTraCk & & & &    \\

%   &ReTraCk+T & & & &  \\
%  \hline 
  % \multirow{4}{*}{A+U}
  % & RnG-KBQA &	91.90 &	73.30 &	81.70 &	47.10 \\
RnG-KBQA + T &	94.30 &	75.90 &	85.90 &	59.50 \\
% & TIARA &	93.76 &	75.22 &	86.35 &	50.84 \\
TIARA + T &	95.10 &	77.77 &	87.86 &	56.88 \\
% & Pangu &	21.40 &	12.17 &	22.32 &	15.32 \\
Pangu + T &	80.51 &	57.52 &	80.15 &	52.85 \\
% & \sys &	64.59 &	36.43 &	65.44 &	43.40 \\
\sys{} + T &	{\bf97.01} &	{\bf89.94} &	{\bf88.31} &	{\bf75.22} \\

  \hline 
\end{tabular}
%}
\end{center}
\caption{Performance of different models for IID and zero-shot test scenarios for unanswerable questions in GrailQAbility for A+U training. Names have the same meanings as in Table~\ref{tab:mainresult}.}
\label{tab:test_scenarios}
%\vspace{-0.5cm}
\end{table}

Additionally, In Table ~\ref{tab:zeroshot_drilldown}, we further scrutinize the performance of the models for sub-categories of zero-shot unanswerable questions. 
We observe that RnG-KBQA and Pangu perform better in terms of both EM and F1(R) for these categories. 
However, this is a natural benefit of mostly predicting $L=NK$ for unanswerable questions, which results in extremely poor performance for the missing data element category, as we have already seen in Table~\ref{tab:drop_type}. As seen earlier, here too all the baseline models have huge variations in performance across full z-shot and partial z-shot categories. 
This is in contrast to \sys{}, which achieves comparable performance. 
It does so without compromising on performance for missing data elements category. 
%with comparatively much less delta in performance across full z-shot and partial z-shot categories. 
This further establishes the stability of \sys{} across various generalization categories and makes it more reliable for real-world scenarios.
%A similar result is obtained on these types of generalization for unanswerable questions (see Table \ref{tab:test_scenarios} in appendix).

%Comment about overall drop in A performance after A+U training.

\begin{table}[h!]
\begin{center}
\small            %\resizebox{1\linewidth}{!}{
\begin{tabular}[b]{|c|cc|cc|}
%\toprule
%\textbf{Memory} & \textbf{Approach} & \multicolumn{5}{c}{\textbf{Datasets}} \\ 
\hline
  \multicolumn{1}{|c|}{Model} & \multicolumn{2}{c|}{Full Z-Shot }  & \multicolumn{2}{c|}{Partial Z-Shot}  \\

& F1(R) & EM & F1(R) & EM \\ \hline
  % & RnG-KBQA &	87.20 &	75.90 &	78.00 &	40.00 \\
RnG-KBQA + T &	89.70 &	86.70 &	{\bf83.10} &	71.00 \\
% & TIARA &	90.15 &	68.97 &	80.25 &	40.45 \\
TIARA + T &	{\bf90.64} &	78.82 &	82.64 &	54.14 \\
% & Pangu &	24.63 &	20.69 &	21.18 &	15.76 \\
Pangu + T &	89.66 &	{\bf89.66} &	79.94 &	{\bf79.46} \\
% & \sys &	57.64 &	25.12 &	43.47 &	11.31 \\
\sys{} + T &	88.67 &	77.83 &	80.89 &	70.54 \\
  \hline 
\end{tabular}
%}
\end{center}
\caption{Performance of different models for missing schema elements - partial zero-shot and full-zero test scenarios in GrailQAbility for A+U training.
Names have the same meanings as in Table~\ref{tab:mainresult}.}
\label{tab:zeroshot_drilldown}
%\vspace{-0.5cm}
\end{table}

 \begin{table*}[h!]
\begin{center}
\small            %\resizebox{1\linewidth}{!}{
\begin{tabular}[b]{|l|rr|rr|rr|rr|}
%\toprule
%\textbf{Memory} & \textbf{Approach} & \multicolumn{5}{c}{\textbf{Datasets}} \\ 
\hline
  \multicolumn{1}{|c|}{Model}&\multicolumn{2}{c|}{Overall}&\multicolumn{2}{c|}{IID}
  
  & \multicolumn{2}{c|}{Compositional}  & \multicolumn{2}{c|}{Zero-Shot}  \\

& F1 & EM & F1 & EM & F1 & EM & F1 & EM \\ \hline
RnG-KBQA &	76.80 &	71.40 & 89.01 &	86.70 &	68.90 &	61.70 &	74.70 &	68.80 \\ 
TIARA & 81.90 & 75.30 & {\bf91.20} & 88.40 & 74.80 & 66.40 & 80.70 & 73.30 \\
Pangu & 82.16 & 75.90 & 86.38 & 81.73 & 76.12 & 68.82 & {\bf82.82} & {\bf76.29 }\\
\sys{}' & {\bf83.33} & {\bf77.84} & {\bf91.22} & {\bf88.58} & {\bf77.49} & {\bf70.48} & 82.32 & 76.20 \\

  \hline 
\end{tabular}
%}
\end{center}
\caption{Performance of different models on GrailQA (validation set) for IID, compositional, and zero-shot test scenarios. 
%Note that we beat previous SOTA on GrailQA. 
\sys{}' denotes \sys{} with EGC}
\label{tab:grailqa_main_table}
%\vspace{-0.5cm}
\end{table*}

\subsection{Results for Answerable-only KBQA}
We designed \sys{} for stability across answerable and unanswerable questions, and this naturally trades off performance across the two categories.
However, we observed that \sys{} performed the best for answerable questions as well in GrailQAbility. 
Motivated by this, we next address research question (3), by evaluating it for traditional KBQA benchmarks with only answerable questions.
Since all questions are answerable in this setting, we (re-)introduce Execution Guided Check (EGC) as the final step for all models including \sys{}. 
With EGC, models output the highest-ranked logical form which when executed over the KB returns a non-empty answer.

In  Table~\ref{tab:grailqa_main_table}, we report results on GrailQA. 
%then logical form is accepted else we keep applying the check until we find a query with non-empty answer. 
%Note that EGC is not applied for GrailQAbility.
%(A training, answerable columns of Table \ref{tab:mainresult} also test the same setting for GrailQAbility). 
We find that overall \sys{} beats previous the state-of-the art by around 1.2 pct points in terms of F1 and 1.8 pct points in terms of EM, establishing a new state-of-the-art for this dataset.
%on GrailQA, and about 3 points on GrailQAbility(A). %This validates that \sys{} is robust across both answerable and unanswerable questions. 
%Further analysis for types of generalization (Table \ref{tab:grailqa_main_table}) shows %very similar trends as for answerable questions in GrailQAbility.
We also see that, as for answerable questions in GrailQAbility, here too \sys{} performs the best for IID and compositional generalization, and performs almost at par with Pangu for zero-shot generalization.
%suggests that \sys{} beats previous best results for IID and compositional generalisation, but for zero-shot generalisation, \sys\ has a comparable or slightly worse performance than Pangu. It is mainly because of traversal dependence tradeoff, as explained in Section~\ref{subsec:ablations}. 
%A similar result is obtained on these types of generalization for unanswerable questions (see Table \ref{tab:test_scenarios} in appendix).

Further analysis shows that \sys{} performs well across questions of various complexities. 
It is the best model for 1, 2, and 4-hop questions, while it is outperformed by TIARA for 3-hop questions. %Table~\ref{tab:num_rels_analysis}.
More details are in the Appendix (~\ref{subsubsec:complexity_analysis_appendix}).

In Table~\ref{tab:webqsp}, we record results for WebQSP.
Here, \sys{} outperforms Pangu by 0.6 pct points but ranks below TIARA by 0.2 pct points. These numbers further establish the usefulness of its architecture for answerable-only KBQA as well.

\begin{table}[h!]
\begin{center}
\small    
\begin{tabular}[b]{|l|r|}
\hline
Model & F1 \\
\hline
TIARA & {\bf75.80} \\
Pangu & 75.00 \\
\sys{}' & 75.60 \\
\hline

\end{tabular}
%}
\end{center}
\caption{Performance of different models on WebQSP (test set) containing only IID answerable questions. We use the WebQSP evaluation script that only reports F1. \sys{}' = \sys{} + EGC}
\label{tab:webqsp}
%\vspace{-0.5cm}
\end{table}

%\section{Analysis} 

% \begin{enumerate}
%     \item Discuss about intuitions and how different modules adds complementary strengths - basic details like coverage plotswith and without different modules.
%     \item Discuss about various trends and trade offs (motivate next step towards robust KGQA) (Comparison across models)
%     \item Error Analysis (Error distribution of the current model)
%     \item Any other ablations ?
%     \item Kind of AUC ROC plots with thresholding - Our method will have better AUC (will highlight strengths on handling the trade-off).
    
% \end{enumerate}

\subsection{Ablation Study} 
\label{subsec:ablations}

%The last three rows of Table \ref{tab:mainresult} establish the contribution at the aggregate level of the different components of \sys{}'s on GrailQAbility for answerable and unanswerable questions.
Finally, we address the research question (4).
Here, we assess the individual contributions of the different components in \sys{}. 
First, we remove (one at a time) the three key components: the logical form integrator (LFI), the logical form retriever (LFR), and the coupled sketch generator (SG) and schema retriever (SR). 
%Note that the sketch generator (SG) and the schema retriever (SR) need to be removed together.
%If we remove schema retriever (SR), then sketch generator (SG) cannot contribute complete logical forms, hence those components have to be ablated together. 
%as all components adds complimentary strengths to different types of answerable and unanswerable questions.

%To  validate our hypotheses about the specific types of questions where each component benefits, we perform addition ablations. 

%To evaluate the benefits of each component, 
Our ablation study focuses on specific question categories.
%In our first experiment, we study coverage analysis (whether correct logical form is in candidates) for unanswerable questions with data element drops. 
First, we study the recall of the correct logical form within the candidate set for unanswerable questions for missing data elements (see Table \ref{tab:dp_sr_ablation_data_drop} in Appendix).
% If we remove SR and SG, the resulting \sys{} ablation only retrieves candidate logical forms via path traversal and hence is not able to enumerate logical forms when paths are incomplete. 
We see that if we remove SR and SG, there is a significant 65 pct point decrease in recall. 
In contrast, removing LFR has little impact. 
This is because LFR relies on path traversal to retrieve candidate logical forms 
This fails when paths are incomplete, preventing \sys{} from effectively enumerating logical forms.
%These happen because if relevant data is not there in the KB, correct data paths are unavailable, and any traversals will be forced to enumerate incorrect paths yielding all incorrect logical forms. 
% \sout{This agrees with our intuition that when relevant data is missing, traversal necessarily retrieves irrelevant logical forms.}
This agrees with our intuition that when essential data elements are missing, we need retrieval techniques that are independent of path traversal.
%We conclude that traversal independent LF construction is necessary for data drop unanswerabilities.

%In a complimentary ablation, we assess the value of traversal-dependent LF construction. For this, we perform the same coverage analysis, but for answerable questions in GrailQAbility. 
Next, we do a similar study of recall 
%of the correct logical form within the candidate set 
for answerable questions (Table \ref{tab:dp_sr_ablation_ans} in Appendix).
We observe that eliminating LFR leads to a significant reduction in recall for zero-shot questions. 
This is unlike removing SG and SR, which only affects i.i.d. questions. This is because retrieval techniques that do not rely on path traversal have a larger search space, resulting in more errors when encountering unseen schema elements. In contrast, path traversal-based methods are grounded at the fact level. 
This results in a smaller search space and therefore fewer errors for zero-shot questions. 
Thus, path-traversal-dependent logical form retrieval is important for zero-shot generalization of answerable questions.
% Removing LFR (and also SR+SG) results in a substantial drop in recall (Table \ref{tab:dp_sr_ablation_ans} in Appendix). 
% Also, LFR has significantly impact for the zero-shot generalization subset of answerable questions. 
% For question forms unseen during training, KB-traversal is the only reliable approach for retrieving logical forms.

%LF retriever is found to be particularly valuable.  
%This happens because traversal independent methods lack KB grounding information and hence make mistakes in selecting unseen schema elements for zero-shot questions. 
%We conclude that traversal-based logical forms are valuable for zero-shot generalization. 

%Finally, we recognize that the space of possible logical forms can be large, which can cause the reranker to make mistakes during discrimination. \sys{} mitigates this via LF integrator, which prunes out invalid logical forms before training. For answerable questions, there exists a second approach. Execution guided constraints (EGC) invoke the next logical form, in case the previous one returns an empty answer. 
Finally, we evaluate the impact of LFI and EGC. 
Both reduce the space of logical form candidates for the discriminator, by pruning out invalid logical forms. Table~\ref{tab:grailqa_lf1_ablation} in the Appendix records performance of ablations of \sys{} in the answerable setting on GrailQA.
%We test their cumulative impact on GrailQA. 
By switching off LFI and EGC separately, we see drops in performance by about 4 pct points and 2 pct points respectively. 
However, on switching off both together, we observe a 17 pct point drop. 
This shows that these components can compensate for each other, but \sys{} needs at least one of them for good performance.
%or else reranker gets confused due to the large space of candidate LFs 

Additional ablations %over questions of different complexities 
show that SG and SR are more helpful for complex (3-hop and 4-hop) questions.
See Sec.~\ref{subsubsec:complexity_analysis_appendix} for more details.

\subsection{Error Analysis}
\label{sec:erroranalysis}

%As A+U training with the thresholding is the most robust setting for \sys{}, we now perform its error analysis on the entire GrailQAbility dataset. 
We have done detailed error analysis for \sys{} on GrailQAbility.
Here, we present a brief summary of it.
% More details are in Sec.~\ref{sec:erroranalysis_appendix}.
For this, we used the the best version \sys{}, which is \sys{}+T with A+U training.
We found three main error categories: (1) \emph{thresholding error}, where, due to thresholding, \sys{} incorrectly predicts NK for a question with a valid logical form; (2) \emph{reranking error}, where the discriminator makes a mistake in scoring, though the candidates contain the correct logical form, and (3) \emph{recall error}, where the correct logical form is not in the set of discriminator candidates. 
This may be due to errors in entity linking, logical form retrieval or logical form construction.
%even fed into the discriminator due to mistakes in earlier stages, like sketch generation, entity linking, schema retrieval or logical form enumeration.

On the subset of answerable questions, thresholding and reranking errors occur for around $38\%$, and $30\%$ of the total errors respectively. 
Improving discriminator calibration can potentially reduce these errors.
The most frequent error is recall error ($70\%$). 
Among these, entity linking errors occur $80\%$ of the time. 
This clearly suggests that improving entity linkers can significantly improve the overall performance of KBQA models.
Unsurprisingly, the majority of the errors across categories occur for the zero-shot generalization questions.
%We also observe that all kinds of errors occur majorly with zero-shot generalization questions. 
%This is not that surprising since relations/mentions not seen at train time may confuse various parts of the model, thereby missing correct logical form in candidates or outputting the correct logical form with a confidence lower than the answerability threshold.
Detailed statistics are in Table \ref{tab:component_wise_errors_k_k} in the Appendix.

For unanswerable questions, we first look at those with missing data elements (Table~\ref{tab:component_wise_errors_k_nk}).
% On the data element drop subset of unanswerable questions, 
We find that the vast majority of errors (around $90\%$) are recall errors, out of which about $72\%$ are attributed to the entity linker. 
Next, thresholding error accounts for $45\%$ of the total errors.
Reranking errors only occur in $5\%$ of total error. 
This suggests that while the ranking of the logical forms according to discriminator assigned scores is correct, the absolute scores have errors.
%the discriminator scores result in the correct ranking of logical forms, but errors exist in the absolute values of scores assigned to the logical forms.

% For unanswerable questions with missing data elements, (around $90\%)$ of errors  are recall errors, out of which about $72\%$ are attributable to the entity linker,  $45\%$ to thresholding and $5\%$ to reranking. 
%This suggests that the discriminator is calibrated well for relative ranking of logical forms, but still errors in assigning correct absolute scores to logical forms.
% See Table~\ref{tab:component_wise_errors_k_nk} for more details.

Finally, we look at the subset of unanswerable questions with missing schema elements.
Since for these the gold logical form is NK, thresholding error is the only source of error. 
This occurs for only $14\%$ of the questions (under the same category), out of which $90\%$ of errors occur for zero-shot generalization. 
This suggests that \sys{} broadly performs well in this setting but 
the scores assigned to logical forms with unseen schema elements have occasional errors.
\section{Conclusions}
\label{sec:conclusion}

We have presented \sys{}, the first specialized KBQA model that shows robust performance for both answerable and unanswerable questions.
To address this challenging task, \sys{} identifies candidate logical forms using data-traversal-based retrieval, as well as schema-based generation via sketch-filling that bridges over data gaps that break traversal.
\sys{} also discriminate between fully formed candidate logical forms at the final stage instead of generating these.
This enables it to better differentiate between valid and invalid logical forms.
In doing so, \sys{} unifies key aspects of different existing KBQA models that assume answerability in IID and transfer settings. 

%To demonstrate this robustness, 

Unlike superficial adaptations of existing SoTA models for unanswerability, we show that \sys{} demonstrates stable performance across adaptation strategies, question categories (answerable and different categories of unanswerable questions) and different generalization settings.
%that performs well for both answerable and unanswerable questions on the GrailQAbility dataset, with and without unanswerability training. 
%Unanswerability performance improves with thresholding and unanswerability training, and it comes with minimal drop in performance for answerable questions.
%Performance is stable across IID, zero-shot and compositional splits for answerable questions, as well as IID and zero-shot splits for unanswerable questions. 
%By comparing against state-of-the-art KBQA models adapted extrinsically for answerability, 
We also show that \sys{} performs significantly better for unanswerable questions and almost at par for answerable ones.
\sys{} also performs well with only answerable training, which is a likely real-world scenario.
\sys{} also retains this stability and performance for answerable-only KBQA benchmarks, 
%Remarkably, \sys{} also 
achieving a new state-of-the-art performance on the answerable-only GrailQA dataset.
% \textcolor{red}{Finally, we conducted a detailed ablation study and demonstrated the importance of different model components for various categories of unanswerability. The \sys{} model architecture can be viewed as a strong framework for robust KBQA models wherein each component (entity linking, complimentary-retrieval techniques, and discriminator) can be individually improved.}
%, demonstrating the strengths of its architecture for KBQA in general.
We release our code-base \footnote{\url{https://github.com/dair-iitd/RetinaQA}} for further research.
%A robust KGQA model is expected to have following characteristics:
%\begin{enumerate}
%    \item It should be robust against answerability detection and performance for both answerable and unanswerable questions should be good (with or without training with unasnwerable questions)
%    \item Gain in performance for unanswerable questions should come with minimal or no drop in performance of answerable questions i.e. A+U training or threshold should not affect answerable performance.
%    \item Performance across different generalizations i.e. IID, Compositional and Zero-shot should be equivalently good for both answerable questions.
%    \item Performance across different generalizations i.e. IID and Zero-shot or Partial Zero-shot and Full Zero-shot should be equivalently good for unanswerable questions.
%\end{enumerate}

%(Can we relate/refer above points in the explanantion below ?)
\section*{Limitations}\label{sec:limitations} 

A sketch, while free of references to the KB, still specifies the length of the path to be traversed in the KB.
The subsequent grounding step is limited by this and cannot adapt the path length after retrieving schema elements from the KB.
\sys{} inherits this limitation from existing sketch generation approaches~\cite{cao:acl2022progxfer,ravishankar:emnlp2022}.
We hope to improve this in future work.

For unanswerable questions without valid logical forms for the given KB, \sys{} only outputs $l=$NK.
However, this does not explain the gap in the schema, which, if bridged, would have make this question answerable. 
The situation is similar for unanswerable questions with valid logical forms but missing data elements.
This is also an important area of future work.

\section*{Risks}
Our work does not have any obvious risks.

\section*{Acknowledgements}
%\textcolor{red}{
Prayushi is supported by a grant from Reliance Foundation. Mausam is supported by a contract with TCS, grants from IBM, Wipro, Verisk, and the Jai Gupta chair fellowship by IIT Delhi. We thank the IIT Delhi HPC facility for its computational resources.
%}

% Entries for the entire Anthology, followed by custom entries
\bibliography{anthology,custom}
\bibliographystyle{acl_natbib}

\appendix
\section{Appendix}

\begin{table*}[h!]
\begin{center}
\small    
\begin{tabular}[b]{|c|c|c|c|c|}
\hline
Components & Overall & IID  & Compositional & Zero-shot \\
\hline
\#questions & 6808 & 3386 & 981 & 2441 \\
total\_errors & 1691 & 445 & 347 & 899 \\
\hline
thresholding\_error & 637 & 161 & 113 & 363 \\
reranking\_error & 508 & 49 & 134 & 325 \\
recall\_error & 1183 & 396 & 213 & 574 \\
\hline
entity\_linking\_error & 949 & 343 & 136 & 470 \\
schema\_retriever\_error & 460 & 61 & 77 & 322 \\
sketch\_parser\_error & 420 & 43 & 154 & 22 \\
\hline

\end{tabular}
%}
\end{center}
\caption{Component wise errors of \sys{} + T (A+U) for answerable questions. Note that reranking errors and recall errors are non-intersecting, while thresholding errors are a subset of total errors i.e. union of reranking and recall errors. Also, recall error is the union of entity\_linking\_error, schema\_retriever\_error, and sketch\_parser\_error while individually these are intersecting.}
\label{tab:component_wise_errors_k_k}
%\vspace{-0.5cm}
\end{table*}

\begin{table}[ht!]
\begin{center}
\small    
\begin{tabular}[b]{|c|c|c|c|}
\hline
Components & Overall & IID  & Zero-shot \\
\hline
\#questions & 1196 & 530 & 666 \\
total\_errors & 287 & 127 & 160 \\
\hline
thresholding\_error & 131 & 59 & 72 \\
reranking\_error & 16 & 5 & 11 \\
recall\_error & 271 & 122 & 149 \\
\hline
entity\_linking\_error & 195 & 77 & 118 \\
schema\_retriever\_error & 56 & 29 & 27 \\
sketch\_parser\_error & 49 & 27 & 22 \\
\hline

\end{tabular}
%}
\end{center}
\caption{Component wise errors of \sys + T (A+U) for data element missing unanswerable questions. Names have the same meanings as in Table ~\ref{tab:component_wise_errors_k_k} }
\label{tab:component_wise_errors_k_nk}
%\vspace{-0.5cm}
\end{table}

\begin{table*}[ht!]
\begin{center}
\small    
\begin{tabular}[b]{|c|cc|cc|cc|cc|}
\hline
\multicolumn{1}{|c}{Model}&\multicolumn{2}{|c}{Overall}&\multicolumn{2}{|c}{IID}
  
  & \multicolumn{2}{|c}{Compositional}  & \multicolumn{2}{|c|}{Zero-Shot}  \\ 
 & F1 & EM & F1 & EM & F1 & EM & F1 & EM  \\ \hline
\sys{}' & 83.33 & 77.84 & 91.22 & 88.58 & 77.49 & 70.48 & 82.32 & 76.2 \\
\sys{}' - EGC & 80.62 & 75.68 & 89.81 & 87.7 & 74.78 & 68.1 & 79.03 & 73.58 \\ 
\sys{}' - LFI  & 78.65 & 73.1 & 88.1 & 84.81 & 75 & 67.31 & 76.04 & 70.4 \\
\sys{}' - LFR  & 71.8 & 68.33 & 87.33 & 85.56 & 69 & 63.47 & 66.19 & 62.83 \\
\sys{}' - (SG $\cup$ SR) & 73.2 & 66.78 & 77.06 & 72.13 & 60.63 & 54.43 & 76.73 & 69.56 \\
\sys{}' - LFI - EGC & 63.29 & 59.99 & 79.84 & 77.84 & 59.33 & 54.03 & 57.73 & 54.68 \\
\hline

\end{tabular}
%}
\end{center}
\caption{Ablation experiment on GrailQA  dev set. EGC refers to Execution Guided Check and LFI refers to Logical Form Integrator, \sys{}' = \sys{} + EGC}
\label{tab:grailqa_lf1_ablation}
%\vspace{-0.5cm}
\end{table*}

\begin{table*}[ht!]
\begin{center}
\small    
\begin{tabular}[b]{|c|c|c|c|c|}
\hline
Model & Overall & IID  & Compositional & Zero-shot \\
\hline
\sys & 82.62 & 88.3 & 78.29 & 76.49 \\
\sys - LFR & 74.24 & 85.91 & 67.38 & 60.79 \\
\sys - (SG $\cup$ SR) &  71.94 & 74.22 & 65.24 & 71.49 \\
\hline

\end{tabular}
%}
\end{center}
\caption{Ablation experiment of Logical Form Recall(\%) on GrailQAbility test set for Answerable questions. LFR refers to Logical Form Retriever, SG refers to Sketch Generation and SR refers to Schema Retriever.}
\label{tab:dp_sr_ablation_ans}
%\vspace{-0.5cm}
\end{table*}

\begin{table}[ht!]
\begin{center}
\small    
\begin{tabular}[b]{|c|c|c|c|}
\hline
Model & Overall & IID  & Zero-shot \\
\hline
\sys & 77.34 & 76.98 & 77.63 \\
\sys - LFR & 77.17 & 76.79 & 77.48 \\
\sys - (SG $\cup$ SR) & 12.29 & 10 & 14.11 \\
\hline

\end{tabular}
%}
\end{center}
\caption{Ablation experiment of Logical Form Coverage(\%) on GrailQAbility test set (Unanswerable questions - Data Element Missing). LFR refers to Logical Form Retriever, SG refers to Sketch Generation and SR refers to Schema Retriever.}
\label{tab:dp_sr_ablation_data_drop}
%\vspace{-0.5cm}
\end{table}

\begin{table}[ht!]
\begin{center}
\small    
\begin{tabular}[b]{ccccc}
\hline
\#relation & 1 & 2  & 3 & 4 \\
\hline
Pangu & 82.8 & 63.5 & 24.7 & 0.0 \\
TIARA & 81.2 & 64.7 & {\bf29.3} & {\bf50.0} \\
RetinaQA & {\bf83.7} & {\bf68.5} & 26.9 & {\bf50.0} \\
\hline
RetinaQA - LFR & 76.3 & 50.9 & 25.1 & 50.0 \\
RetinaQA - (SP $\cup$ SR) & 72.9 & 56.9 & 12.0 & 0.0 \\
\hline

\end{tabular}
%}
\end{center}
\caption{Performance for different types of questions on the GrailQA validation set in terms of EM. \#relation
denotes the number of relations in the s-expression.}
\label{tab:num_rels_analysis}
%\vspace{-0.5cm}
\end{table}

\subsection{Entity Linker}
\label{subsec:appendix_entity_linker}
We use an off-the-shelf entity linker~\cite{ye:acl2022rngkbqa} previously used in the KBQA literature~\cite{shu-etal-2022-tiara,gu-etal-2023-dont}, which uses a standard 3-staged pipeline - Mention Detection, Candidate Generation, and Entity Disambiguation.  
Mention Detector first identifies span of text from question which corresponds to name of an entity. For each mention a set of candidates entities are generated using alias mapping of FACC1~\cite{facc1}. Final stage is a neural disambiguator which rank candidates given the question and context of entities.

\subsection{Implementation Details}
\label{subsec:implementation_details}
To perform experiments for GrailQAbility, we first update the original Freebase KG using codebase\footnote{\url{https://github.com/dair-iitd/GrailQAbility}}. To test baselines for GrailQAbility, we use the existing codebases\footnote{\url{https://github.com/dki-lab/Pangu}} \footnote{\url{https://github.com/microsoft/KC/tree/main/papers/TIARA}} \footnote{\url{https://github.com/salesforce/rng-kbqa}} and make changes in code to adapt for answer-ability detection. All of the baselines assumes answerability and employs Execution Guide Check i.e. if a logical form returns an empty answer upon execution then they select next best logical form. We have removed this constraint while performing experiments for GrailQAbility. Also for A+U training we have made code changes so that models can be trained to predict logical form as $NK$ unanswerable questions.
We implement our model using Pytorch ~\cite{pytorch} and Hugging Face\footnote{\url{https://huggingface.co/}}. All the experiments of \sys{} are performed using an NVIDIA A100 GPU with 80 GB RAM. Above mentioned configurations are the maximum ones, since we have different components and all do not require same compute configurations. For Sketch Generation we fine tune Seq2Seq t5-base model for 10 epochs (fixed). We use learning rate of $3\text{e-}5$ and batch size of $8$. We use beam search during decoding with $beam size = 10$. We also check syntactic correctness while selecting top ranked sketch. For WebQSP sketch generator is trained for $15$ epochs with batch size of $2$ Training time for sketch parser is around 3 hours. LF Integrator is a parameter free module and does not require any training. Since, LF Integrator converts logical forms into query-graphs and validates type-level constraints, it is a costly operation. So we employ parallel processing(with cache) for this stage i.e. we use 4-6 CPUs (each with 2 cores) to create pool of valid logical forms. It takes around 5 hours to generate candidates for all train, dev and test data. 
Finally we train Discriminator which fine-tune t5-base Seq2Seq model. We train Discriminator with learning rate $1\text{e-}4$ and batch size 4 for $10$ epochs. For discriminator training we use AdmaW ~\cite{loshchilov2019decoupled} optimizer and linear scheduler with warm up ratio of $0.01$. We use 64 negative samples per question for contrastive training. Generally discriminator model converges in 2 epochs of training so we use patience of 2 i.e. if best model does not change for consequent 2 epochs then we assume model has converged and will stop training. It takes around 7-8 hours to train a discriminator. We train WebQSP for $15$ epochs with patience equal to $10$ and $32$ negative samples. Inference time for discriminator is few minutes.

For A+U training components like Entity Linker, Schema Retriever, LF Retriever are trained only on question where logical form is known. While training for questions with $l=$"NK" is performed only at last step.

All the results presented for single run (however the reproducibility of results is already verified). 
We release our code-base\footnote{\url{https://anonymous.4open.science/r/RETINAQA-122B}} for the community.

% \subsection{Dataset and Eval Metrics}
% \begin{enumerate}
%     \item Define IID, comp, zero-shot and other definitions (unanswerabilties), EGC.
%     \item details about F1 (L), F1(R)
% (atleast add citation. ~\cite{patidar-etal-2023-knowledge}
% \end{enumerate}

% \subsection{Analysis}
% Analysis across various parameters. (We have results ready)
% \begin{enumerate}
%     \item Functions/ Operators
%     \item \# rels (hops)
% \end{enumerate}

\subsection{In Depth Analysis} \label{subsec:analysis_appendix}
\subsubsection{Trade-off Analysis} \label{subsec:trade-off}
Sec~\ref{subsec:ablations} describes how individual components strengthens performance for different types of answerabilties and unanswerabilties. This section discusses an important trade-off  i.e. %This section will discuss two improtant tarde-off analysis that highlights the complimentary nature of individual components and how they interact with each other. 
\textbf{Traversal dependent Retrieval Vs Traversal independent Retrieval} : Traversal based Retrieval methods perform step by step enumeration over KB to retrieve next possible set of candidates(which is retrieval at data level). While Traversal independent Retrieval based method generate candidates based on semantic similarity with the question(which is at schema level). So for Data Element Missing unanswerability where data paths are missing, Traversal based methods will never find correct path during enumeration and hence will not be able to reach to a correct logical form. While Traversal independent method can generate correct logical form. Hence Traversal independent methods performs well for data element missing.\\
At the same time the search space for Traversal independent methods is much larger as it lacks KB grounding information. So for zero-shot generalisation where schema elements are unseen Traversal dependent tends to get confused between similar schema elements.

\subsubsection{Complexity Analysis} \label{subsubsec:complexity_analysis_appendix}

Tab.~\ref{tab:num_rels_analysis} records the performance of RetinaQA for queries of different complexities represented by number of relations in s-expression (or number of hops in answer path). We can see that for 1-hop and 2-hop questions RetinaQA is better than both baselines, while for 3-hop questions RetinaQA is not the best but is better than Pangu. Further by comparing ablations i.e. without Logical Form Retriever and without (Sketch Generation and Schema Retriever) we can see that Sketch Generation and Schema Retriever contribute more to the performance of 3-hop and 4-hop questions.

\subsection{GrailQAbility - Dataset Creation}
\label{subsec:appendix_dataset_creation}
We summarise the dataset creation algorithm of GrailQAbility~\cite{patidar-etal-2023-knowledge} here. In a nutshell, the authors start with a standard KBQA dataset containing only answerable questions for a given KB. Then they introduce unanswerability in steps, by deleting schema elements (entity types and relations) and data elements (entities and facts) from the given KB. They mark questions that become unanswerable as a result of each deletion with appropriate unanswerability labels. So starting from a larger set of all answerable questions, the authors create two subsets of data - one set of answerable questions and another set of unanswerable questions (which are unanswerable due to missing structures in graph/KB).

\end{document}